%% file: main.tex
\documentclass[lettersize,journal]{IEEEtran}
\usepackage{amsmath,amsfonts}
\usepackage{algorithmic}
\usepackage{algorithm}
\usepackage{array}
\usepackage[caption=false,font=normalsize,labelfont=sf,textfont=sf]{subfig}
\usepackage{textcomp}
\usepackage{stfloats}
\usepackage{url}
\usepackage{verbatim}
\usepackage{graphicx}
\usepackage{cite}
\usepackage{orcidlink}
\usepackage{booktabs}
\usepackage{multirow}

\usepackage{microtype}
\usepackage{amsmath}
\usepackage{threeparttable}

\begin{document}

\title{SPECTRA: State-Space Exogenous Context and Temporal-Frequency Resolution Architecture for Probabilistic Energy Forecasting}

\author{Hang Ye, Xinyan Jiang, Yuedong Shi, Yangxin Zhu, Jianming Wei, Tian Zheng, Xiaoying Zheng, Yongxin Zhu

\thanks{This work was supported in part by the {National Natural Science Foundation of China (Grant No. 12373113 and No. 12475196)}.(\textit{Corresponding author: Xiaoying Zheng})

Hang Ye and Xinyan Jiang are with Shanghai Advanced Research Institute, Chinese Academy of Sciences, Shanghai 201210, China, and also with the University of Chinese Academy of Sciences, Beijing 101408, China (e-mail: yehang@sari.ac.cn; jiangxy2024@sari.ac.cn).

Yuedong Shi are with School of Physical Science and Technology, Shanghaitech University, Shanghai 201210, China (e-mail: shiyd@shanghaitech.edu.cn)

Tian Zheng are with State Power Investment Corporation Limited, Beijing 100029, China (e-mail: zhengtian@snerdi.com.cn)

Yangxin Zhu are with China University of Petroleum, Beijing 102249, China (e-mail: 2025215063@cup.edu.cn)

Jianming Wei,
Xiaoying Zheng and Yongxin Zhu are with Shanghai Advanced Research Institute, Chinese Academy of Sciences, Shanghai 201210, China (e-mail: wjm@sari.ac.cn; zhengxy@sari.ac.cn; zhuyongxin@sari.ac.cn).

    The relevant code can be found at this repository: \url{https://github.com/yedadasd/SPECTRA}

}}

\markboth{Submitted to IEEE Transactions on Power Systems}%
{Shell \MakeLowercase{\textit{et al.}}: A Sample Article Using IEEEtran.cls for IEEE Journals}


\maketitle

\begin{abstract}
Modern power systems increasingly require probabilistic forecasts amid interacting uncertainties from renewable intermittency, flexible demand, market volatility, and weather-dependent generation. However, existing methods often treat multi-scale decomposition, exogenous-variable alignment, and probabilistic output as separate steps, obscuring how predictable structures and uncertainty-bearing fluctuations jointly shape the forecast distribution. This paper proposes a state-space exogenous-context and temporal-frequency resolution architecture for general probabilistic energy forecasting. Its central premise is that trend-periodic components primarily determine the baseline trajectory, whereas high-frequency residuals and external perturbations govern the spread and asymmetry of forecast uncertainty. Accordingly, the architecture adaptively separates deterministic and residual streams, aligns exogenous context with both, refines the deterministic backbone through multi-resolution spectral-temporal state-space modeling, and estimates ordered quantile boundaries from their complementary representations. Experiments on load, price, solar, and wind forecasting achieve the best continuous ranked probability score in 14 of 18 settings, reducing average CRPS by 5.74\% and upper-tail quantile risk by 7.27\% over the strongest baselines. These results support deterministic-stochastic separation as an effective design principle for general probabilistic energy forecasting.

\end{abstract}


\input{sections/1_introduction}
\input{sections/2_preliminaries}
\input{sections/3_method}
\input{sections/4_experiments}

\input{sections/5_conclusion}

\section*{Acknowledgment}

The authors gratefully acknowledge the Shanghai Academy of Artificial Intelligence for Science (SAIS) and the Power Dispatching Control Center of China Southern Power Grid for providing the NewEnergy2025 dataset through the New Energy Track of the 3rd World Science Intelligence Competition.

\bibliographystyle{IEEEtran}
\bibliography{references}

\end{document}

%% file: sections/1_introduction.tex
\section{Introduction}
\label{sec:intro}

\IEEEPARstart{A}{c}curate energy forecasting underpins the secure and economic operation of modern power systems, supporting unit commitment, economic dispatch, demand response, and electricity market trading~\cite{Hong2020EnergyForecastingReview, Wang2023DiffLoadUQ}. This task is increasingly challenging as both sides of the power system evolve. On the demand side, data centers and large-scale artificial intelligence computing clusters are emerging as large, spatially concentrated, and partially flexible loads whose rapid growth amplifies demand variability~\cite{Chen2025ElectricityDA}. On the supply side, increasing wind and photovoltaic penetration introduces stronger intermittency and non-stationarity. Consequently, modern energy forecasting is no longer merely a point-estimation problem for load, price, wind, or solar power, but a probabilistic modeling task that must characterize temporal dynamics, external perturbations, and the risks associated with extreme fluctuations~\cite{HONG2016914}.

The central difficulty arises from three intertwined properties of energy time series. First, their dynamics are inherently multi-scale: daily and weekly operating patterns coexist with long-term trends, seasonal effects, and short-lived spikes. Second, they are driven by heterogeneous weather, calendar, market, and system-status variables whose effects vary across targets and temporal components. Third, power-system decisions require calibrated uncertainty rather than just a forecast center, especially under renewable ramps, price spikes, and volatile flexible loads. These properties suggest that a general probabilistic forecaster should organize multi-scale temporal structure, exogenous context, and stochastic boundaries within a unified modeling process rather than treating them independently.

Existing methods address parts of this problem, but rarely in a unified manner. Classical statistical, physical, and shallow machine learning approaches provide interpretable baselines and often incorporate domain variables explicitly. Wavelet-ARIMA models, for example, describe electricity price volatility~\cite{Conejo2005WaveletARIMA}, while semi-parametric additive models combine calendar variables, lagged load, and temperature for load forecasting with prediction intervals~\cite{Fan2012Load}. Related statistical, physical, and hybrid designs have also been applied to solar and wind forecasting, where meteorological conditions and system attributes are decisive~\cite{Pedro2012NoExogenous, Foley2012WindReview}. However, their reliance on handcrafted features and task-specific assumptions limits their ability to capture nonlinear interactions, long-range dependencies, and sudden regime shifts in modern energy systems.

Deep learning improves representation capacity, but many models primarily strengthen the sequence backbone rather than the complete probabilistic forecasting pipeline. Recurrent and convolutional networks have been applied to fluctuating residential load, probabilistic autoregressive forecasting, and photovoltaic power prediction~\cite{Kong2019LSTMLoad, xiong2021attention, salinas2020deepar, Zang2018PVHybridCNN}. Transformer-based models further capture long-range dependencies through self-attention. Autoformer integrates decomposition to model trend and periodic patterns, whereas PatchTST uses patch-based tokenization~\cite{wu2021autoformer, nie2023patchtst}. Nevertheless, self-attention incurs quadratic computational and memory costs with input length and may overfit small, noisy, or non-stationary energy datasets. State-space models such as Mamba offer a promising linear-complexity alternative with input-dependent dynamics~\cite{gu2023mamba, ye2025karma}. However, using them only as generic sequence backbones does not directly resolve the energy-specific requirements of temporal decomposition, component-wise exogenous alignment, and uncertainty estimation.

Decomposition-based modeling provides another important line of evidence. Because energy series contain regular operating patterns and volatile residual fluctuations, many methods separate temporal components before or during prediction. Classical time-frequency techniques, including wavelet transform, variational mode decomposition, and wavelet-packet analysis, have been widely used~\cite{Conejo2005WaveletARIMA, he2019hybrid, zhang2020adaptive, liu2023deep}. Recent neural models introduce learnable or moving-average decomposition, including Autoformer, DLinear, and the energy-specific probabilistic model TADNet~\cite{wu2021autoformer, zeng2023dlinear, ye2024tadnet}. These studies demonstrate that separating trend-like structures from residual fluctuations can improve forecasting. However, existing decomposition designs are commonly coupled with deterministic point forecasting or task-specific probabilistic heads, while their interaction with exogenous-variable alignment and uncertainty-boundary estimation remains insufficiently unified. Consequently, how temporal components and external drivers jointly shape predictive uncertainty is often unclear.

Exogenous variables and probabilistic outputs have also been investigated, but external information is often treated as an auxiliary input rather than explicitly aligned with distinct temporal components. Temperature and calendar effects are critical for load forecasting~\cite{Fan2012Load}; numerical weather prediction, weather scenarios, and Bayesian model averaging improve solar uncertainty estimation~\cite{huang2016semi, sun2020probabilistic, zhang2025weather}; and multi-source temporal attention exploits heterogeneous information for wind forecasting~\cite{zhang2021multi}. Temporal Fusion Transformer combines variable selection, gating, and interpretable attention for multi-horizon forecasting with static, known-future, and historical covariates~\cite{lim2021temporal}. DeepAR and adaptive probabilistic load forecasting emphasize uncertainty-aware prediction~\cite{salinas2020deepar, de2023adaptive}, while TimeXer models endogenous-exogenous interactions through variate tokens and cross-attention~\cite{wang2024timexer}. Despite these advances, existing methods commonly rely on input concatenation, global fusion, or generic probabilistic heads. Such designs may obscure how external drivers differently affect predictable temporal structures and uncertainty-bearing residual fluctuations.

The above discussion indicates that the key challenge is not merely to design a stronger sequence model. A general energy forecaster should distinguish predictable operating structure from uncertainty-bearing residual variation, align heterogeneous exogenous drivers with both components, and produce calibrated probabilistic boundaries. Motivated by this view, we introduce a \textbf{S}tate-s\textbf{P}ace \textbf{E}xogenous \textbf{C}ontext and \textbf{T}emporal-Frequency \textbf{R}esolution \textbf{A}rchitecture (SPECTRA) for general probabilistic energy forecasting. The Macro-Trend \& Periodic Decoupling module (MTPD) first separates the series into a trend-periodic backbone and a high-frequency residual. In parallel, the Exogenous Context Synergizer (ECS) treats exogenous variables as contextual tokens and aligns them with both streams, allowing external drivers to affect baseline dynamics and residual volatility differently. The Spectral-Temporal State-Space Engine (STSSE) refines the deterministic backbone through multi-resolution spectral-temporal representation and linear-time state-space propagation. Finally, the Stochastic Boundary Estimator (SBE) integrates deterministic and residual features to generate ordered quantile forecasts and prediction intervals. The main contributions are summarized as follows:
\begin{itemize}
    \item \textbf{deterministic-stochastic decoupling paradigm.}
    We separate predictable trend-periodic structures from uncertainty-bearing residuals, allowing the forecast center and probabilistic boundaries to follow dedicated but coupled pathways. MTPD performs adaptive temporal-frequency decomposition, while SBE estimates ordered quantiles from deterministic and residual features without assuming a fixed distribution.
    \item \textbf{Component-aware context and dependency modeling.}
    Building on this decoupling, ECS aligns exogenous variables separately with the two streams, enabling differentiated effects on baseline dynamics and residual volatility. STSSE further captures multi-resolution and long-range dependencies in the deterministic stream through wavelet analysis and linear-time state-space modeling.
    \item \textbf{A unified framework with cross-domain validation.}
    We integrate the four modules into an end-to-end framework for probabilistic load, price, solar, and wind forecasting. Experiments on ECL, OPS, and GEFCom2014 show that SPECTRA achieves the best CRPS in 14 of 18 settings and ranks within the top two in 17, reducing the average CRPS by 5.74\% and upper-tail quantile risk $\rho_{90}$ by 7.27\% over the strongest baseline. These results validate deterministic--stochastic decoupling as an effective design principle for general probabilistic energy forecasting.
\end{itemize}

The rest of the paper is organized as follows. Section~\ref{sec:preliminaries} introduces the problem formulation and preliminaries. Section~\ref{sec:method} presents SPECTRA. Section~\ref{sec:experiments} reports benchmark, ablation, and model-analysis results. Section~\ref{sec:conclusion} concludes the paper and outlines future work.

%% file: sections/2_preliminaries.tex
\section{Preliminaries}
\label{sec:preliminaries}

This section establishes the problem formulation and theoretical foundations of SPECTRA, including probabilistic energy forecasting, quantile regression and uncertainty evaluation, and the state-space modeling used in STSSE.

\subsection{Problem Statement}
\label{sec:preliminaries:problem}

Consider a general probabilistic energy forecasting task covering electricity load, market price, wind power, and photovoltaic generation. Let $\mathbf{y}_t \in \mathbb{R}^{D_y}$ denote the vector of $D_y$ endogenous target variables at time step $t$, and let $\mathbf{x}_t \in \mathbb{R}^{D_x}$ represent multi-source exogenous covariates, such as meteorological conditions and temporal markers.

Given a look-back window of length $L$, the historical endogenous sequence is
$\mathbf{X}=[\mathbf{y}_{t-L+1},\dots,\mathbf{y}_t]^T \in \mathbb{R}^{L\times D_y}$.
The exogenous context covering both the historical and future horizons is defined as
$\mathbf{X}_{exo}=[\mathbf{x}_{t-L+1},\dots,\mathbf{x}_{t+H}]^T \in \mathbb{R}^{(L+H)\times D_x}$.

Unlike deterministic forecasting, which produces a single point estimate, probabilistic forecasting estimates the conditional quantiles of the future trajectory
$\mathbf{Y}=[\mathbf{y}_{t+1},\dots,\mathbf{y}_{t+H}]^T \in \mathbb{R}^{H\times D_y}$.
Given $Q$ target probability levels $\mathcal{Q}=\{\tau_1,\tau_2,\dots,\tau_Q\}$, where $\tau_q\in(0,1)$, the predicted quantile tensor is
\begin{equation}
    \hat{\mathbf{Y}}=\mathcal{F}\left(\mathbf{X},\mathbf{X}_{exo};\mathbf{\Theta}\right)\in \mathbb{R}^{H\times Q\times D_y},
\end{equation}
where $\mathcal{F}(\cdot)$ denotes a forecasting model parameterized by $\mathbf{\Theta}$. The parameters $\mathbf{\Theta}$ are optimized to minimize quantile-consistent errors between the predicted tensor $\hat{\mathbf{Y}}$ and the observed future trajectory $\mathbf{Y}$, as specified in Section~\ref{sec:preliminaries:quantile}. For each probability level $\tau_q$, each entry $\hat{y}_{h,q,d}$ satisfies
$\mathbb{P}(y_{t+h,d}\leq\hat{y}_{h,q,d}\mid\mathbf{X},\mathbf{X}_{exo})=\tau_q$, for $h=1,\dots,H$ and $d=1,\dots,D_y$.

\subsection{Quantile Regression and Uncertainty Evaluation}
\label{sec:preliminaries:quantile}

We adopt quantile regression to obtain distribution-free uncertainty estimates without assuming a parametric error distribution~\cite{xu2023quantile}. For a quantile level $\tau\in\mathcal{Q}$, the pinball loss for prediction error $e=y-\hat{y}$ is
\begin{equation}
    \rho_{\tau}(e)=e\left(\tau-\mathbb{I}_{\{e<0\}}\right)=\max\left(\tau e,(\tau-1)e\right),
    \label{eq:pinball_loss}
\end{equation}
where $\mathbb{I}_{\{\cdot\}}$ is the indicator function. The loss over all forecast steps, target channels, and quantile levels is
\begin{equation}
\label{eq:quantile_loss}
    \mathcal{L}_{QR}=\frac{1}{H D_y Q}\sum_{t=1}^{H}\sum_{d=1}^{D_y}\sum_{q=1}^{Q}\rho_{\tau_q}\left(y_{t,d}-\hat{y}_{t,q,d}\right),
\end{equation}
where $\hat{y}_{t,q,d}$ is the prediction for channel $d$, horizon step $t$, and quantile $\tau_q$. Minimizing $\mathcal{L}_{QR}$ jointly learns the quantile boundaries that characterize the predictive distribution.

Probabilistic performance is evaluated by the Continuous Ranked Probability Score (CRPS), which measures calibration and sharpness~\cite{berrisch2023crps}. Given a predictive cumulative distribution function $\hat{F}(z)$ and observation $y$,
\begin{equation}
    \label{eq:crps}
    \mathrm{CRPS}(\hat{F},y)=\int_{-\infty}^{\infty}\left(\hat{F}(z)-\mathbb{I}_{\{y\leq z\}}\right)^2\,\mathrm{d}z.
\end{equation}
For a predictive distribution represented by finite quantiles, CRPS is approximated by
\begin{equation}
    \label{eq:discrete_crps}
    \mathrm{CRPS}(y,\hat{\mathbf{y}})\approx\frac{2}{Q}\sum_{q=1}^{Q}\rho_{\tau_q}\left(y-\hat{y}^{(\tau_q)}\right).
\end{equation}
This approximation links quantile-based training with probabilistic evaluation.

\subsection{State-Space Model Foundations}
\label{sec:preliminaries:ssm}
State-Space Models (SSMs) efficiently model sequences by propagating a compact latent state. Classical SSMs are widely used in control and time-series estimation, with the Kalman Filter as a representative example~\cite{kalman1960new}. Deep SSMs extend this formulation to neural sequence modeling, mapping an input $x(t) \in \mathbb{R}$ to a latent state $h(t) \in \mathbb{R}^{N}$ and output $y(t) \in \mathbb{R}$:
\begin{equation}
    \dot{h}(t) = \mathbf{A}h(t) + \mathbf{B}x(t),
\end{equation}
\begin{equation}
    y(t) = \mathbf{C}h(t) + \mathbf{D}x(t),
\end{equation}
where $\mathbf{A} \in \mathbb{R}^{N \times N}$, $\mathbf{B} \in \mathbb{R}^{N \times 1}$, $\mathbf{C} \in \mathbb{R}^{1 \times N}$, and $\mathbf{D} \in \mathbb{R}^{1 \times 1}$ are the transition, input, output, and direct-transmission parameters.

For discrete time series, the continuous system is discretized with step size $\Delta \in \mathbb{R}^{+}$. Under the Zero-Order Hold (ZOH) assumption,
\begin{equation}
    \bar{\mathbf{A}} = \exp(\Delta \mathbf{A}),
\end{equation}
\begin{equation}
    \bar{\mathbf{B}} = (\Delta \mathbf{A})^{-1}
    \left(
        \exp(\Delta \mathbf{A}) - \mathbf{I}
    \right)
    \Delta \mathbf{B}.
\end{equation}
The resulting recurrence becomes
\begin{equation}
    h_t = \bar{\mathbf{A}}h_{t-1}+\bar{\mathbf{B}}x_t,
\end{equation}
\begin{equation}
    y_t=\mathbf{C}h_t+\mathbf{D}x_t.
\end{equation}

Modern selective SSMs make $\Delta$, $\mathbf{B}$, and $\mathbf{C}$ input-dependent, adapting temporal dynamics to each sequence~\cite{gu2021efficiently, gu2023mamba}. This property enables STSSE to model long-range seasonal patterns and short-term variations in the deterministic backbone without the quadratic cost of full self-attention.



%% file: sections/3_method.tex
\section{Proposed Method: SPECTRA}
\label{sec:method}

\begin{figure*}[t]
  \centering
  \includegraphics[width=\linewidth]{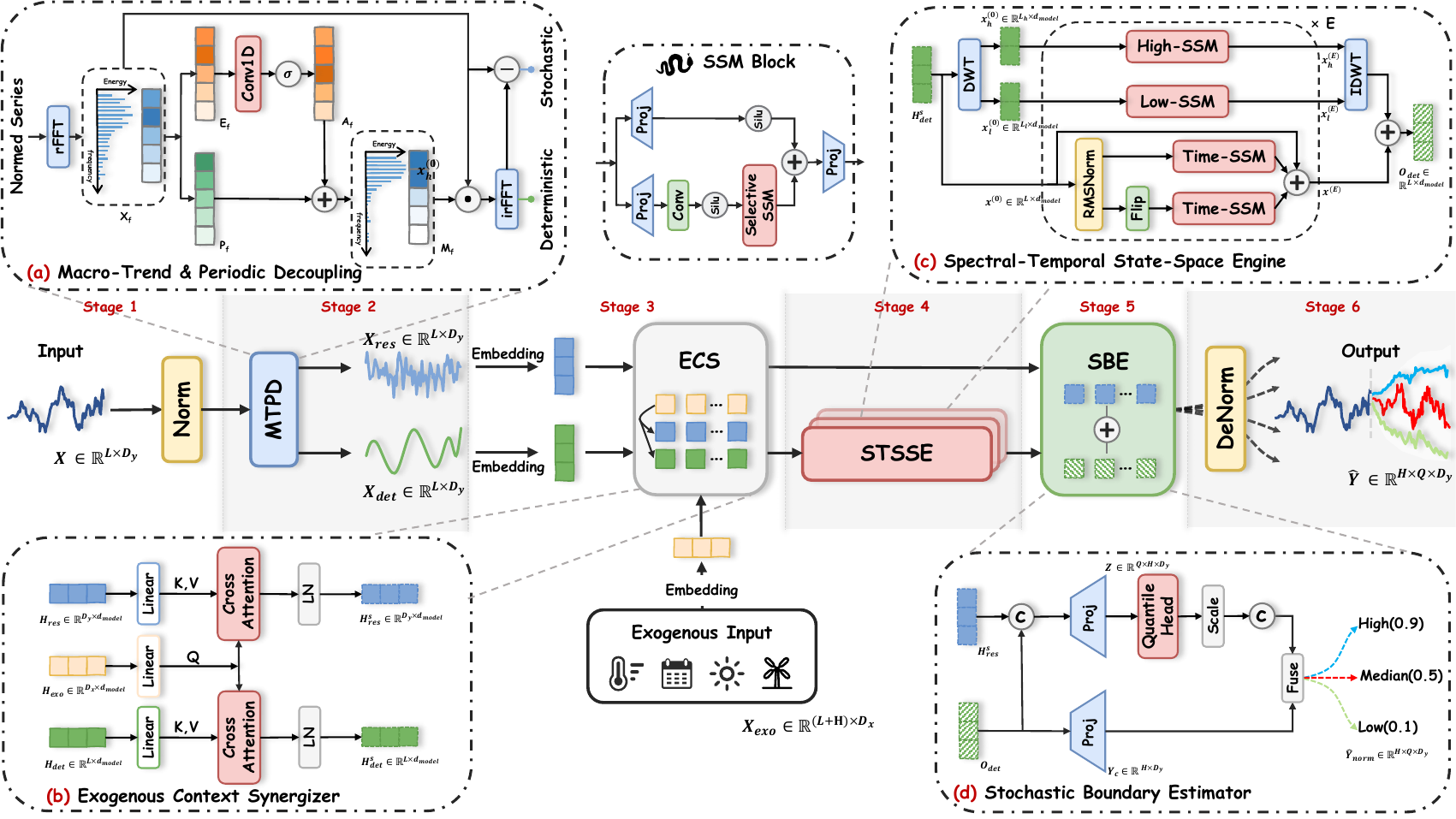}
  \caption{Architecture of SPECTRA. The main pipeline contains six stages
        from normalization to quantile-aware denormalization, while insets
        (a)--(d) detail MTPD, ECS, STSSE, and SBE, respectively. The
        deterministic and residual streams follow specialized paths and are
        fused for quantile forecasting.}
  \label{fig:spectra-architecture}
\end{figure*}

The design of SPECTRA follows a deterministic-stochastic decoupling principle. In energy systems, regular operating cycles, seasonal variations, and slowly varying market states mainly determine the forecast center, whereas renewable ramps, price spikes, weather perturbations, and flexible-load disturbances mainly affect the spread and asymmetry of future uncertainty. If these patterns are modeled through a single latent pathway, the forecast center and uncertainty-bearing fluctuations may be entangled, making probabilistic outputs difficult to interpret and calibrate.

Based on this observation, SPECTRA follows the six-stage pipeline in Fig.~\ref{fig:spectra-architecture}, centered on four coupled modules. MTPD separates the normalized endogenous sequence into a trend-periodic backbone and a high-frequency residual; ECS aligns exogenous context with both streams; STSSE selectively refines the context-conditioned deterministic backbone through multi-resolution filtering and state-space propagation; and SBE fuses the refined deterministic and residual features to estimate quantile boundaries. This asymmetric routing assigns each component a mechanism consistent with its statistical role: structured patterns are modeled by the state-space backbone, whereas volatile residuals remain directly available for boundary estimation. Consequently, it reduces cross-component interference, makes the source of predictive uncertainty more explicit, and preserves efficient long-range modeling.

Given a historical endogenous window $\mathbf{X} \in \mathbb{R}^{L \times D_y}$ and a multi-source exogenous context sequence $\mathbf{X}_{exo} \in \mathbb{R}^{(L+H) \times D_x}$, SPECTRA outputs a quantile tensor $\hat{\mathbf{Y}} \in \mathbb{R}^{H \times Q \times D_y}$, as defined in Section~\ref{sec:preliminaries:problem}. The overall forward process is summarized below:

\textbf{Stage 1: Instance Normalization.} To reduce non-stationary scale shifts across samples, SPECTRA applies reversible instance Normalization~\cite{kim2021reversible} to each endogenous input window:
\begin{equation}
    \mathbf{X}_{norm} =
    \frac{\mathbf{X}-\boldsymbol{\mu}}
    {\sqrt{\boldsymbol{\sigma}^2+\epsilon}}
    \in \mathbb{R}^{L \times D_y},
\end{equation}
where $\boldsymbol{\mu} \in \mathbb{R}^{D_y}$ and $\boldsymbol{\sigma}^2 \in \mathbb{R}^{D_y}$ are the mean and variance computed along the look-back dimension, and $\epsilon$ is a small constant.

\textbf{Stage 2: Adaptive Temporal-Frequency Decoupling.} MTPD applies an adaptive spectral mask to decompose the normalized series into a deterministic backbone and a residual stream:
\begin{equation}
    \mathbf{X}_{det},\mathbf{X}_{res}=\mathrm{MTPD}(\mathbf{X}_{norm}),
\end{equation}
where $\mathbf{X}_{det},\mathbf{X}_{res}\in\mathbb{R}^{L\times D_y}$. The deterministic stream retains dominant trend-periodic structures, while the residual stream preserves high-frequency deviations that are important for uncertainty estimation.

\textbf{Stage 3: Exogenous Context Alignment.} The deterministic stream is embedded as temporal tokens, while the residual stream retains trajectory-level variate tokens:
\begin{equation}
\label{eq:embedded}
\begin{split}
    \mathbf{H}_{det} &= \mathrm{Embedding}_{det}(\mathbf{X}_{det}), \\
    \mathbf{H}_{res} &= \mathrm{Embedding}_{res}(\mathbf{X}_{res}),
\end{split}
\end{equation}
where $\mathbf{H}_{det}\in\mathbb{R}^{L\times d_{model}},\mathbf{H}_{res}\in\mathbb{R}^{D_y\times d_{model}}$. The exogenous sequence, which may include both historical and known future variables, is embedded over the joint time span:
\begin{equation}
    \mathbf{H}_{exo}
    =
    \mathrm{Dropout}\left(\mathbf{X}_{exo}^{T}\mathbf{W}_{exo}+\mathbf{b}_{exo}\right)
    \in \mathbb{R}^{D_x\times d_{model}},
\end{equation}
where $\mathbf{W}_{exo}\in\mathbb{R}^{(L+H)\times d_{model}}$. ECS produces context-conditioned deterministic and residual representations:
\begin{equation}
    \mathbf{H}_{det}^{s},\mathbf{H}_{res}^{s}=\mathrm{ECS}
    \left(
    \mathbf{H}_{det},
    \mathbf{H}_{res},
    \mathbf{H}_{exo}
    \right).
\end{equation}

\textbf{Stage 4: State-Space Backbone Refinement.} The context-conditioned deterministic stream is processed by STSSE:
\begin{equation}
    \mathbf{O}_{det}
    =
    \mathrm{STSSE}(\mathbf{H}_{det}^{s})
    \in \mathbb{R}^{L\times d_{model}}.
\end{equation}
This stage strengthens long-range dependency modeling in the predictable backbone without the quadratic cost of full self-attention.

\textbf{Stage 5: Quantile Boundary Estimation.} SBE aggregates the refined deterministic feature $\mathbf{O}_{det}$ and the context-conditioned residual feature $\mathbf{H}_{res}^{s}$ to estimate dynamic quantile boundaries. In parallel, a linear shortcut maps the decoupled backbone to a stable horizon-level baseline:
\begin{equation}
\label{eq:shortcut}
    \mathbf{Y}_{base} = \mathrm{Linear}_{base}(\mathbf{X}_{det}^T) \in \mathbb{R}^{H \times D_y},
\end{equation}
The normalized probabilistic output is then
\begin{equation}
    \hat{\mathbf{Y}}_{norm}= \mathrm{Permute}\left(\mathrm{SBE}(\mathbf{O}_{det}, \mathbf{H}_{res}^{s})\right) + \mathbf{Y}_{base} \in \mathbb{R}^{H \times Q \times D_y}.
\end{equation}

\textbf{Stage 6: Quantile-Aware Denormalization.} The predicted quantiles are restored to the original scale by broadcasting the instance statistics along the horizon and quantile dimensions:
\begin{equation}
    \hat{\mathbf{Y}}^{(\tau_q)} = \sqrt{\boldsymbol{\sigma}^2+\epsilon}\,\hat{\mathbf{Y}}_{norm}^{(\tau_q)} + \boldsymbol{\mu} \in \mathbb{R}^{H \times D_y}.
\end{equation}
Here, $\hat{\mathbf{Y}}^{(\tau_q)}$ is the final denormalized forecast at the $q$-th quantile level. The full architecture is trained end-to-end with the objective in Section~\ref{sec:method:training}.

Stages 1-6 complete the SPECTRA pipeline through normalization, temporal-frequency decoupling, exogenous context alignment, state-space refinement, quantile estimation. The four core modules are detailed next, beginning with MTPD.

\subsection{Macro-Trend \& Periodic Decoupling}
\label{sec:method:mtpd}

Energy series contain slowly varying operating structures and short-lived perturbations. The former provides a stable basis for the forecast center, while the latter carries much of the uncertainty associated with ramps, spikes, and abrupt external changes. MTPD explicitly separates these two parts in the frequency domain.

For the $d$-th target variate, the $k$-th real Fourier coefficient of the normalized input  $\mathbf{X}_{norm} \in \mathbb{R}^{L \times D_y}$ is
\begin{equation} 
    \begin{aligned} 
        \mathbf{X}_{f}(k,d) &= \sum_{t=0}^{L-1} \mathbf{X}_{norm}(t,d) \exp\left(-j\frac{2\pi kt}{L}\right), \\ k&=0,\ldots,F-1,\quad d=0,\ldots,D_y-1, 
    \end{aligned} 
\end{equation} 
where $F=\lfloor L/2 \rfloor+1$ is the number of non-redundant frequency bins. In compact form,
\begin{equation} 
    \mathbf{X}_{f} = \mathrm{rFFT}(\mathbf{X}_{norm}) \in \mathbb{C}^{F \times D_y}. 
\end{equation} 

To construct an input-adaptive spectral filter, MTPD first computes the log-amplitude energy:
\begin{equation} 
    \mathbf{E}_{f} = \log\left(1+|\mathbf{X}_{f}|\right) \in \mathbb{R}^{F \times D_y},
\end{equation} 
An adaptive mask is generated by a lightweight convolution along the frequency axis:
\begin{equation} 
    \mathbf{A}_{f} = \sigma\left(\mathrm{Conv1D}(\mathbf{E}_{f})\right) \in \mathbb{R}^{F \times D_y},
\end{equation} 
where $\sigma(\cdot)$ is the sigmoid function. This mask emphasizes dominant frequency components according to the spectral structure of each input instance.

Since the forecast center is mainly associated with low-frequency and dominant periodic components, MTPD also introduces a smooth low-frequency prior:
\begin{equation} 
    \mathbf{P}_{f}(k) = \sigma\left(\frac{f_c-k/F}{\tau}\right), \quad k=0,\ldots,F-1, 
\end{equation} 
where $f_c$ is the cutoff frequency and $\tau$ controls the transition sharpness. The final spectral mask combines the prior and the adaptive mask:
\begin{equation} 
    \mathbf{M}_{f} = \sigma(\alpha)\mathbf{P}_{f} + \left(1-\sigma(\alpha)\right)\mathbf{A}_{f}, 
\end{equation} 
where $\alpha$ is a learnable scalar and $\mathbf{P}_{f}$ is broadcast over target variates. The decomposed components are obtained by inverse transformation and subtraction:
\begin{align} 
    \mathbf{X}_{det} &= \mathrm{irFFT}\left(\mathbf{X}_{f}\odot\mathbf{M}_{f}\right) \in \mathbb{R}^{L \times D_y}, \\ 
    \mathbf{X}_{res} &= \mathbf{X}_{norm}-\mathbf{X}_{det}. 
\end{align}

Unlike moving average or pooling decomposition methods, MTPD avoids the use of a fixed temporal window: the prior term preserves stable macro patterns, while the adaptive term enables the decomposition split to adapt to the spectral characteristics of each input instance.

\subsection{Exogenous Context Synergizer}
\label{sec:method:ecs}

Exogenous variables are not merely auxiliary covariates in energy forecasting. Their effects are component-dependent: regular calendar and temperature patterns may reshape the baseline trajectory, while abrupt meteorological or market changes may mainly enlarge residual volatility. ECS is designed to align external context with the decomposed streams rather than simply concatenate all variables at the input layer, in order to better capture the dynamic relationships between external factors and the decomposed components.

Given the embeddings in Eq.~\eqref{eq:embedded}, ECS uses two parallel cross-attention paths~\cite{vaswani2017attention}. The deterministic stream queries the exogenous tokens to extract context relevant to trend-periodic dynamics, and the residual stream queries the same exogenous tokens to capture context related to short-term fluctuations:
{\small
\begin{align}
    \mathbf{C}_{res} &= \mathrm{CrossAtt}_{res}(Q=\mathbf{H}_{res}, K=\mathbf{H}_{exo}, V=\mathbf{H}_{exo}), \\
    \mathbf{C}_{det} &= \mathrm{CrossAtt}_{det}(Q=\mathbf{H}_{det}, K=\mathbf{H}_{exo}, V=\mathbf{H}_{exo}).
\end{align}}
The attended contexts are injected into their corresponding streams through residual connections and layer normalization:
\begin{align}
    \mathbf{H}_{res}^{s} &= \mathrm{LayerNorm}(\mathbf{H}_{res} + \mathrm{Dropout}(\mathbf{C}_{res})), \\
    \mathbf{H}_{det}^{s} &= \mathrm{LayerNorm}(\mathbf{H}_{det} + \mathrm{Dropout}(\mathbf{C}_{det})),
\end{align}
where $\mathbf{H}_{det}^{s}\in \mathbb{R}^{L \times d_{model}},\mathbf{H}_{res}^{s} \in \mathbb{R}^{D_y \times d_{model}}$ are the context-conditioned representations. Compared with direct fusion, ECS selectively routes external context to the decomposed streams while preserving the deterministic-residual separation from MTPD. The deterministic output is passed to STSSE, and the residual output is delivered to SBE.

\subsection{Spectral-Temporal State-Space Engine}
\label{sec:method:stsse}

After MTPD and ECS, $\mathbf{H}_{det}^{s}$ contains the trend-periodic backbone and its aligned exogenous context. However, this representation may still contain latent multi-resolution structures, such as coarse seasonal variations and finer local deviations. STSSE refines the deterministic backbone by combining wavelet-based multi-resolution filtering with selective state-space propagation.

Let
\begin{equation}
    \mathbf{x}^{(0)} = \mathbf{H}_{det}^{s} \in \mathbb{R}^{L \times d_{model}}.
\end{equation}
A one-level discrete wavelet transform~\cite{mallat1989theory} is applied along the temporal axis. For latent channel $r$, the approximation and detail coefficients are
\begin{align}
    \mathbf{x}_{l}^{(0)}(n,r)&=\sum_{t}g[t-2n]\mathbf{x}^{(0)}(t,r),\\
    \mathbf{x}_{h}^{(0)}(n,r)&=\sum_{t}h[t-2n]\mathbf{x}^{(0)}(t,r),
\end{align}
where $g[\cdot]$ and $h[\cdot]$ are low-pass and high-pass wavelet filters. Equivalently,
\begin{equation}
    (\mathbf{x}_{l}^{(0)},\mathbf{x}_{h}^{(0)})=\mathrm{DWT}_{t}(\mathbf{x}^{(0)}),
\end{equation}
where $\mathbf{x}_{l}^{(0)}\in\mathbb{R}^{L_l\times d_{model}}$ and $\mathbf{x}_{h}^{(0)}\in\mathbb{R}^{L_h\times d_{model}}$.

STSSE stacks $E$ blocks. In the $e$-th block, the approximation and detail subbands are processed by two dedicated selective state-space layers~\cite{gu2023mamba}:
\begin{equation}
    \mathbf{x}_{l}^{(e)} = \mathrm{SSM}_{l}^{(e)} \left( \mathbf{x}_{l}^{(e-1)} \right), \quad \mathbf{x}_{h}^{(e)} = \mathrm{SSM}_{h}^{(e)} \left( \mathbf{x}_{h}^{(e-1)} \right).
\end{equation}
This allows coarse and fine deterministic patterns to follow different latent dynamics.

In parallel, the main deterministic stream is refined by bidirectional selective state-space propagation. With RMS normalization~\cite{zhang2019root}, the forward and backward scans are
{\small
\begin{align}
    \mathbf{f}^{(e)} &= \mathrm{SSM}_{\rightarrow}^{(e)} \left( \mathrm{RMSNorm}\left(\mathbf{x}^{(e-1)}\right) \right), \\
    \mathbf{b}^{(e)} &= \mathrm{Flip} \left( \mathrm{SSM}_{\leftarrow}^{(e)} \left( \mathrm{Flip} \left( \mathrm{RMSNorm}\left(\mathbf{x}^{(e-1)}\right) \right) \right) \right),
\end{align}}
where $\mathrm{Flip}(\cdot)$ reverses the scanning order of latent features. The main stream is updated via a residual connection as:
\begin{equation}
    \mathbf{x}^{(e)} = \mathbf{f}^{(e)} + \mathbf{b}^{(e)} + \mathbf{x}^{(e-1)}.
\end{equation}
After $E$ blocks, the refined subbands are reconstructed and fused with the main stream:
\begin{equation}
    \mathbf{O}_{det} = \mathrm{LayerNorm} \left( \mathrm{IDWT} \left( \mathbf{x}_{l}^{(E)}, \mathbf{x}_{h}^{(E)} \right) + \mathbf{x}^{(E)} \right).
\end{equation}
The output $\mathbf{O}_{det}\in\mathbb{R}^{L\times d_{model}}$ then undergoes a linear transformation to convert this temporal representation into $\mathbf{O}_{det}\in\mathbb{R}^{D_y\times d_{model}}$, and serves as the deterministic structural feature for SBE. Since selective state-space propagation scales linearly with sequence length, STSSE enables efficient dependency modeling for long energy histories.

\subsection{Stochastic Boundary Estimator}
\label{sec:method:sbe}

Probabilistic energy forecasts often exhibit input-dependent and asymmetric uncertainty. Instead of assuming a fixed predictive distribution, SBE estimates quantile boundaries from the deterministic structural feature $\mathbf{O}_{det}$ and the context-conditioned residual feature $\mathbf{H}_{res}^{s}$. The former anchors the prediction around the refined trend-periodic backbone, while the latter adjusts the interval width and asymmetry according to residual volatility and exogenous perturbations.

Given $\mathbf{O}_{det},\mathbf{H}_{res}^{s}\in\mathbb{R}^{D_y\times d_{model}}$, SBE first estimates a horizon-level center:
\begin{equation}
    \mathbf{Y}_{c} = \mathrm{Linear}_{c}(\mathbf{O}_{det}) \in \mathbb{R}^{H\times D_y}.
\end{equation}
It then concatenates the deterministic and residual features and maps them into raw quantile-wise offsets:
{\small
\begin{equation}
    \mathbf{Z} = \mathrm{Reshape}\left(\mathrm{Linear}_{q}\left([\mathbf{O}_{det}\,\Vert\,\mathbf{H}_{res}^{s}]\right)\right) \in \mathbb{R}^{Q\times H\times D_y},
\end{equation}}
where $\Vert$ denotes feature concatenation. Let $q_m$ denote the median index satisfying $\tau_{q_m}=0.5$. The raw offsets are transformed into direction-aware stochastic offsets:
{\small
\begin{equation}
    \Delta^{(\tau_q)} = \begin{cases} -\lambda |\mathbf{Z}^{(\tau_q)}|, & q < q_m, \\ \mathbf{Z}^{(\tau_q)}, & q = q_m, \\ \lambda |\mathbf{Z}^{(\tau_q)}|, & q > q_m, \end{cases}
\end{equation}}
where $\lambda>0$ is a learnable or trainable scale parameter. This direction-aware construction places lower and upper quantiles on opposite sides of the median center and reduces cross-side quantile violations.

Combining the MTPD shortcut baseline in Eq.~\eqref{eq:shortcut}, the normalized forecast for the $q$-th quantile is
{\small
\begin{equation}
    \hat{\mathbf{Y}}_{norm}^{(\tau_q)} = \mathbf{Y}_{base} + \underbrace{\big(\mathbf{Y}_{c} + \Delta^{(\tau_q)}\big)}_{\text{SBE output}}, \quad q=1,\ldots,Q. 
\end{equation}}
Stacking all quantiles gives $\hat{\mathbf{Y}}_{norm}\in\mathbb{R}^{H\times Q\times D_y}$, which is then denormalized in Stage 6. This design avoids an explicit noise distribution and uses the residual stream to shape asymmetric predictive boundaries around the deterministic center.

\subsection{Loss Function}
\label{sec:method:training}
Quantile regression calibrates probabilistic boundaries yet fails to retain energy trajectories' dominant periodic structure explicitly. Inspired by~\cite{wang2025fredf}, SPECTRA integrates the quantile regression loss in Eq.~\eqref{eq:quantile_loss} with a lightweight frequency-domain regularizer; the former guarantees temporal calibration, the latter cuts spectral distortion across the forecast horizon.

Let $\mathcal{F}(\cdot)$ denote the rFFT along the temporal dimension, and let $\Omega_K$ be the indices of the top-$K$ frequency components selected from the target amplitude spectrum during training. The spectral amplitude loss is
{\small
\begin{equation}
    \mathcal{L}_{F} = \frac{1}{Q} \sum_{q=1}^{Q} \left\| \left| \mathcal{F}\left(\hat{\mathbf{Y}}^{(\tau_q)}\right) \right|_{\Omega_K} - \left| \mathcal{F}\left(\mathbf{Y}\right) \right|_{\Omega_K} \right\|_{1},
    \label{eq:spectral_loss}
\end{equation}}
where $|\cdot|_{\Omega_K}$ extracts the amplitudes of the selected dominant frequencies. The final training objective is
\begin{equation}
    \mathcal{L} = \mathcal{L}_{QR} + \eta \mathcal{L}_{F}.
    \label{eq:training_objective}
\end{equation}
where $\eta$ controls the contribution of spectral regularization. This objective jointly promotes time-domain quantile calibration and frequency-domain consistency.

%% file: sections/4_experiments.tex
\section{Experiments}
\label{sec:experiments}

This section evaluates SPECTRA for probabilistic forecasting across load, electricity price, solar, and wind tasks. We first compare the model with representative baselines—probabilistic, covariate-aware, Transformer-based, and decomposition-based methods—on public benchmarks. Ablation studies assess each core module's contribution, followed by a case study on renewable energy. Finally, we analyze long-horizon stability and computational efficiency to examine the model's practical accuracy and efficiency trade-off.

\begin{table*}[t]
\centering
\caption{Comprehensive Probabilistic Forecasting Results on ECL, OPS, and GEF Datasets}
\label{tab:main_results}
\resizebox{\textwidth}{!}{%
\begin{tabular}{ll | cc cccccccc cccccccc}
\toprule
\multirow{3}{*}[-1.5ex]{Model} & \multirow{3}{*}[-1.5ex]{Metric} & \multicolumn{2}{c}{ECL} & \multicolumn{8}{c}{OPS} & \multicolumn{8}{c}{GEF} \\
\cmidrule(lr){3-4} \cmidrule(lr){5-12} \cmidrule(lr){13-20}
& & \multicolumn{2}{c}{ } & \multicolumn{2}{c}{Load} & \multicolumn{2}{c}{Price} & \multicolumn{2}{c}{Solar} & \multicolumn{2}{c}{Wind} & \multicolumn{2}{c}{Load} & \multicolumn{2}{c}{Price} & \multicolumn{2}{c}{Solar} & \multicolumn{2}{c}{Wind} \\
\cmidrule(lr){3-4} \cmidrule(lr){5-6} \cmidrule(lr){7-8} \cmidrule(lr){9-10} \cmidrule(lr){11-12} \cmidrule(lr){13-14} \cmidrule(lr){15-16} \cmidrule(lr){17-18} \cmidrule(lr){19-20}
& & $\mathcal{S}$ & $\mathcal{L}$ & $\mathcal{S}$ & $\mathcal{L}$ & $\mathcal{S}$ & $\mathcal{L}$ & $\mathcal{S}$ & $\mathcal{L}$ & $\mathcal{S}$ & $\mathcal{L}$ & $\mathcal{S}$ & $\mathcal{L}$ & $\mathcal{S}$ & $\mathcal{L}$ & $\mathcal{S}$ & $\mathcal{L}$ & $\mathcal{S}$ & $\mathcal{L}$ \\
\midrule
\multirow{3}{*}{DeepAR}
& CRPS            & 0.279 & 0.711 & 0.480 & 0.481 & 0.554 & 0.592 & 0.221 & 0.545 & 0.547 & 0.615 & 0.213 & 0.368 & 0.223 & 0.458 & 0.398 & 0.579 & 0.609 & 0.743 \\
& $\rho_{50}$     & 0.406 & 0.982 & 0.829 & 0.830 & 0.841 & 0.891 & 0.337 & 0.923 & 0.858 & 0.961 & 0.305 & 0.518 & 0.326 & 0.667 & 0.568 & 0.800 & 0.935 & 1.144 \\
& $\rho_{90}$     & 0.231 & 0.581 & 0.307 & 0.299 & 0.460 & 0.492 & 0.184 & 0.497 & 0.471 & 0.548 & 0.159 & 0.259 & 0.201 & 0.441 & 0.374 & 0.548 & 0.448 & 0.527 \\
\midrule
\multirow{3}{*}{Autoformer}
& CRPS            & 0.313 & 0.321 & 0.264 & 0.274 & 0.545 & 0.592 & 0.221 & 0.238 & 0.582 & 0.603 & 0.262 & 0.301 & 0.382 & 0.426 & 0.472 & 0.504 & 0.603 & 0.624 \\
& $\rho_{50}$     & 0.443 & 0.460 & 0.344 & 0.376 & 0.768 & 0.840 & 0.292 & 0.323 & 0.880 & 0.947 & 0.342 & 0.445 & 0.448 & 0.561 & 0.648 & 0.720 & 0.906 & 1.012 \\
& $\rho_{90}$     & 0.271 & 0.276 & 0.241 & 0.236 & 0.485 & 0.566 & 0.202 & 0.211 & 0.553 & 0.560 & 0.195 & 0.213 & 0.433 & 0.449 & 0.439 & 0.461 & 0.451 & 0.419 \\
\midrule
\multirow{3}{*}{TFT}
& CRPS            & 0.219 & 0.239 & 0.178 & 0.205 & 0.408 & 0.473 & 0.192 & 0.203 & 0.508 & 0.644 & \underline{0.172} & 0.279 & 0.202 & 0.329 & 0.366 & 0.451 & \underline{0.462} & 0.634 \\
& $\rho_{50}$     & 0.296 & 0.328 & 0.229 & 0.278 & 0.595 & 0.691 & 0.244 & 0.269 & 0.782 & 0.986 & \textbf{0.254} & 0.429 & 0.257 & 0.466 & 0.522 & 0.654 & \underline{0.718} & 0.993 \\
& $\rho_{90}$     & 0.197 & 0.216 & 0.154 & 0.170 & 0.339 & 0.389 & 0.198 & 0.204 & 0.442 & 0.591 & 0.130 & 0.199 & 0.193 & 0.319 & 0.343 & 0.413 & \underline{0.339} & 0.461 \\
\midrule
\multirow{3}{*}{DLinear}
& CRPS            & 0.231 & 0.249 & 0.202 & 0.235 & 0.392 & 0.462 & 0.221 & 0.233 & 0.412 & 0.544 & 0.224 & 0.281 & 0.239 & 0.345 & 0.398 & 0.472 & 0.464 & 0.582 \\
& $\rho_{50}$     & 0.262 & 0.298 & 0.217 & 0.285 & 0.528 & 0.652 & \underline{0.205} & \underline{0.229} & \underline{0.630} & \textbf{0.849} & 0.278 & 0.405 & 0.278 & 0.478 & 0.534 & 0.648 & \textbf{0.705} & \textbf{0.884} \\
& $\rho_{90}$     & 0.228 & 0.240 & 0.219 & 0.238 & 0.318 & 0.374 & 0.177 & 0.187 & 0.361 & 0.509 & 0.181 & 0.200 & 0.232 & 0.333 & 0.372 & 0.428 & 0.345 & 0.428 \\
\midrule
\multirow{3}{*}{TiDE}
& CRPS            & 0.291 & 0.316 & 0.250 & 0.294 & 0.462 & 0.558 & 0.265 & 0.281 & 0.529 & 0.680 & 0.246 & 0.332 & 0.295 & 0.439 & 0.465 & 0.559 & 0.567 & 0.707 \\
& $\rho_{50}$     & 0.272 & 0.312 & 0.226 & 0.297 & 0.561 & 0.695 & 0.220 & 0.246 & 0.682 & 0.970 & 0.296 & 0.444 & 0.281 & 0.507 & 0.573 & 0.717 & 0.773 & 0.994 \\
& $\rho_{90}$     & 0.320 & 0.346 & 0.276 & 0.315 & 0.435 & 0.525 & 0.337 & 0.355 & 0.515 & 0.634 & 0.217 & 0.268 & 0.335 & 0.484 & 0.456 & 0.542 & 0.462 & 0.558 \\
\midrule
\multirow{3}{*}{PatchTST}
& CRPS            & 0.200 & 0.224 & 0.161 & 0.199 & \underline{0.330} & \textbf{0.419} & 0.181 & 0.198 & \textbf{0.404} & \textbf{0.538} & 0.177 & 0.263 & 0.189 & 0.304 & \underline{0.334} & \underline{0.423} & 0.467 & 0.584 \\
& $\rho_{50}$     & 0.256 & 0.295 & 0.193 & 0.263 & \textbf{0.474} & \textbf{0.603} & 0.233 & 0.269 & \textbf{0.630} & \underline{0.865} & 0.266 & 0.410 & \underline{0.237} & \textbf{0.420} & \textbf{0.471} & \textbf{0.607} & 0.724 & 0.920 \\
& $\rho_{90}$     & 0.187 & 0.208 & 0.147 & 0.172 & \underline{0.274} & \underline{0.348} & 0.180 & 0.189 & \underline{0.351} & \textbf{0.472} & 0.131 & 0.189 & 0.177 & 0.292 & 0.321 & 0.403 & 0.341 & 0.406 \\
\midrule
\multirow{3}{*}{TimeXer}
& CRPS            & 0.193 & 0.218 & 0.160 & 0.195 & 0.343 & 0.431 & 0.196 & 0.207 & 0.435 & 0.556 & 0.174 & \underline{0.254} & 0.196 & 0.319 & 0.354 & 0.426 & 0.473 & 0.571 \\
& $\rho_{50}$     & \underline{0.244} & \underline{0.285} & \underline{0.193} & \underline{0.254} & 0.495 & 0.621 & 0.246 & 0.272 & 0.681 & 0.880 & 0.259 & \underline{0.398} & 0.246 & 0.441 & 0.505 & 0.618 & 0.744 & 0.917 \\
& $\rho_{90}$     & 0.179 & 0.201 & 0.143 & 0.167 & 0.284 & 0.359 & 0.189 & 0.196 & 0.373 & 0.505 & 0.129 & 0.180 & 0.183 & 0.301 & 0.337 & 0.395 & \textbf{0.337} & 0.401 \\
\midrule
\multirow{3}{*}{TADNet}
& CRPS            & \underline{0.169} & \underline{0.197} & \underline{0.134} & \underline{0.184} & 0.395 & 0.487 & \underline{0.146} & \underline{0.156} & 0.417 & \underline{0.540} & 0.186 & 0.255 & \underline{0.189} & \underline{0.304} & 0.340 & 0.426 & 0.478 & \underline{0.562} \\
& $\rho_{50}$     & 0.252 & 0.292 & 0.200 & 0.275 & 0.584 & 0.713 & 0.222 & 0.239 & 0.653 & 0.865 & 0.293 & 0.408 & 0.285 & 0.454 & 0.508 & 0.634 & 0.768 & 0.931 \\
& $\rho_{90}$     & \underline{0.132} & \underline{0.157} & \underline{0.095} & \underline{0.133} & 0.323 & 0.413 & \underline{0.106} & \underline{0.113} & 0.369 & 0.489 & \underline{0.126} & \textbf{0.178} & \underline{0.157} & \underline{0.283} & \underline{0.307} & \underline{0.395} & 0.346 & \textbf{0.378} \\
\midrule
\multirow{3}{*}{\textbf{SPECTRA}}
& CRPS            & \textbf{0.150} & \textbf{0.187} & \textbf{0.107} & \textbf{0.162} & \textbf{0.326} & \underline{0.423} & \textbf{0.127} & \textbf{0.145} & \underline{0.406} & 0.552 & \textbf{0.166} & \textbf{0.253} & \textbf{0.141} & \textbf{0.291} & \textbf{0.326} & \textbf{0.419} & \textbf{0.462} & \textbf{0.558} \\
& $\rho_{50}$     & \textbf{0.225} & \textbf{0.278} & \textbf{0.160} & \textbf{0.241} & \underline{0.484} & \underline{0.619} & \textbf{0.197} & \textbf{0.227} & 0.632 & 0.868 & \underline{0.259} & \textbf{0.397} & \textbf{0.212} & \underline{0.426} & \underline{0.477} & \underline{0.611} & 0.719 & \underline{0.896} \\
& $\rho_{90}$     & \textbf{0.120} & \textbf{0.153} & \textbf{0.076} & \textbf{0.114} & \textbf{0.258} & \textbf{0.344} & \textbf{0.085} & \textbf{0.095} & \textbf{0.348} & \underline{0.486} & \textbf{0.114} & \underline{0.179} & \textbf{0.113} & \textbf{0.258} & \textbf{0.298} & \textbf{0.385} & 0.349 & \underline{0.391} \\
\bottomrule 
\end{tabular}%
}
\par\vspace{1mm} 
\noindent
\begin{minipage}{0.98\textwidth}
\footnotesize
\textbf{Bold} and \underline{underlined} values indicate the best and second-best results, respectively. $\mathcal{S}$ and $\mathcal{L}$ denote average performance over short-term horizons ($12$, $24$, and $36$) and long-term horizons ($72$, $120$, and $168$), respectively. For Wind and Solar, $\mathcal{L}$ uses only the $72$-step horizon due to their high volatility.

\end{minipage}
\end{table*}

\subsection{Experimental Setup}
\label{sec:exp_setup}

\subsubsection{Datasets}
\label{sec:exp_datasets}

We evaluate SPECTRA on three public benchmarks and one recent renewable case-study dataset, covering load, price, solar, and wind forecasting. \textbf{Electricity (ECL)} is derived from the UCI ElectricityLoadDiagrams dataset~\cite{trindade2015electricityloaddiagrams20112014} and records hourly electricity consumption of 321 clients from 2012 to 2014. \textbf{OpenPowerSystem (OPS)} uses hourly Open Power System Data~\cite{wiese2019open} from 2015 to 2017, including 59 load, 31 price, 36 solar, and 57 wind series. \textbf{GEFCom2014 (GEF)}~\cite{hong2016probabilistic} includes probabilistic forecasting tracks for load, price, solar, and wind. Its load and price tracks span 2011-2014 with temperature, calendar, holiday, and load-forecast covariates, while its solar and wind tracks span 2012-2013 with weather covariates. \textbf{NewEnergy2025 (NE)}, which can be accessed in the code repository and is adopted solely in Section~\ref{sec:exp_case_study}, includes normalized 15-minute generation records collected from five wind farms and five photovoltaic plants, together with meteorological forecasts obtained from three adjacent stations spanning the period from 2015 to 2017.

For ECL, OPS, and GEF, we adopt chronological training/validation/test splits with a ratio of 7:1:2, following common long-term forecasting protocols~\cite{wu2021autoformer, nie2023patchtst, wang2024timexer}. Unless otherwise specified, the input length is set to $L=168$ to cover one week of hourly observations. All datasets are normalized using statistics from the training set, and no additional task-specific preprocessing is applied. This ensures a fair comparison among different models.

\subsubsection{Baselines}
\label{sec:exp_baselines}
The compared methods are selected to cover the main modeling paradigms used in modern energy and time-series forecasting, with competitive representatives from each category. DeepAR~\cite{salinas2020deepar} is included as a classical autoregressive probabilistic forecaster, while TFT~\cite{lim2021temporal} and TADNet~\cite{ye2024tadnet} represent covariate-aware and energy-oriented probabilistic architectures. To assess whether SPECTRA improves beyond recent general-purpose forecasting backbones, we further compare it with TiDE~\cite{das2023tide}, TimeXer~\cite{wang2024timexer}, Autoformer~\cite{wu2021autoformer}, and PatchTST~\cite{nie2023patchtst}, which cover MLP-based, exogenous-variable, decomposition-enhanced, and patch-based Transformer designs. DLinear~\cite{zeng2023dlinear} is also included as a lightweight decomposition-based baseline to examine whether the proposed modules bring gains beyond simple linear trend-seasonality modeling. 

For originally deterministic models, we replace only the final prediction layer with a quantile projection head, training them with the same quantile regression objective as SPECTRA. All methods produce identical quantile sets for consistent probabilistic evaluation.

\subsubsection{Evaluation Metrics}
\label{sec:exp_metrics}

We evaluate forecasting performance using both deterministic and probabilistic metrics. 
For deterministic accuracy, the median quantile is used as the point forecast, and the mean squared error is computed as
\begin{equation}
    \mathrm{MSE} = \frac{1}{NHD_y}\sum_{n=1}^{N}\sum_{h=1}^{H}\sum_{d=1}^{D_y} \left(y_{n,h,d} - \hat{y}_{n,h,d}^{(0.5)}\right)^2,
\end{equation}
where $N$, $H$, and $D_y$ denote the number of test samples, forecasting horizon, and target-variable dimension, respectively.

For probabilistic evaluation, all models output quantiles corresponding to $\mathcal{Q} = \{0.01, 0.02, \ldots, 0.99\}$. We report the normalized quantile risk at representative quantile levels.
\begin{equation}
    \rho_{\tau} = \frac{2\sum_{n,h,d} \ell_{\tau}\left(y_{n,h,d} - \hat{y}_{n,h,d}^{(\tau)}\right)}{\sum_{n,h,d} |y_{n,h,d}|},
\end{equation}
where $\ell_{\tau}(\cdot)$ is the pinball loss defined in Section~\ref{sec:preliminaries:quantile}. 
Here, $\rho_{50}$ reflects the normalized median forecasting error, while $\rho_{90}$ emphasizes upper-tail risk, which is important for reserve scheduling, price-spike anticipation, and renewable ramping events.

CRPS is computed using the quantile-score approximation in Eq.~\eqref{eq:discrete_crps} and aggregated over all valid test entries:
\begin{equation}
    \mathrm{CRPS} = \frac{2\sum_{\tau\in\mathcal{Q}}\sum_{(n,h,d)\in\mathcal{V}}\ell_{\tau}\!\left(y_{n,h,d}-\hat{y}_{n,h,d}^{(\tau)}\right)}{|\mathcal{Q}|\sum_{(n,h,d)\in\mathcal{V}}|y_{n,h,d}|} .
\end{equation}
where $\mathcal{V}=\{(n,h,d):y_{n,h,d}\neq 0\}$ denotes the valid evaluation entries. 
Lower values indicate better performance for all reported metrics.

\subsubsection{Implementation Details}
\label{sec:exp_implementation}


All models are implemented in PyTorch and trained under the same data split, prediction horizons, and evaluation protocol. We adopt the Adam optimizer with a learning rate of $5\times10^{-4}$, a batch size of 32, a maximum of 10 training epochs, and an early stopping strategy with a patience of 3. Experimental results are averaged over three random seeds: $\{2024,2025,2026\}$. For the SPECTRA model, the configuration parameters are set as follows: $d_{model}=512$, the number of STSSE layers is $E=3$, the core selective state-space module uses an expansion factor of 1, a state dimension of $d_{state}=16$, and a local convolution width of $d_{conv}=2$. All experiments are conducted on a single NVIDIA A30 GPU.

\begin{figure*}[t]
    \centering
    \includegraphics[width=\textwidth]{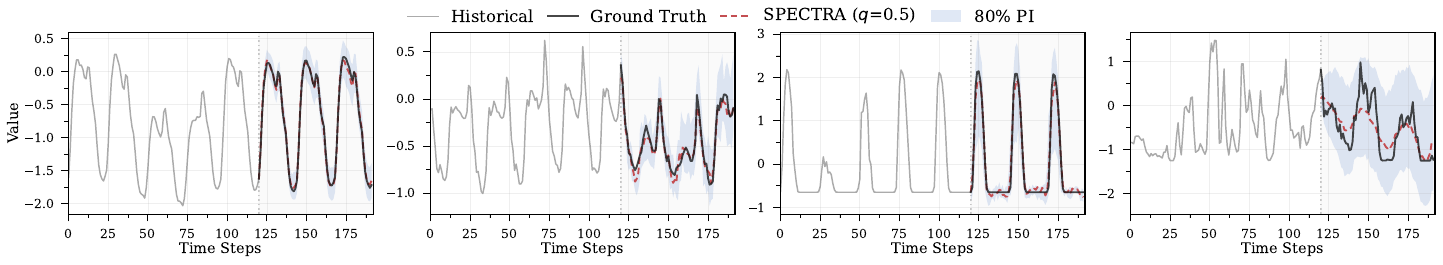}
    \includegraphics[width=\textwidth]{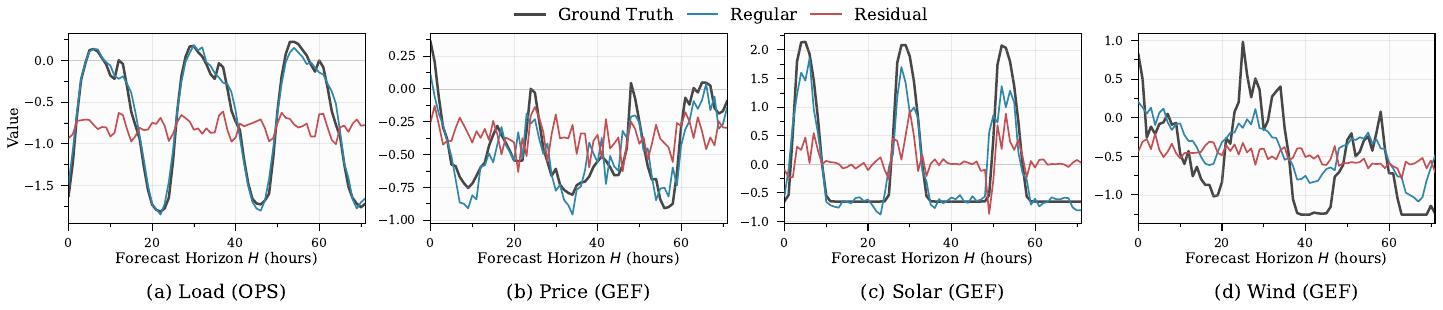}
    \caption{Forecasting results of SPECTRA with $H=72$: forecasts and 80\% prediction intervals (top), and deterministic and residual components (bottom).}
    \label{fig:result_visualization}
\end{figure*}

\subsection{Main Results}
\label{sec:exp_main_results}

Table~\ref{tab:main_results} reports the results on ECL, OPS, and GEF across load, price, solar, and wind forecasting tasks. SPECTRA achieves the best CRPS in 14 of 18 settings and ranks within the top two in 17, reducing the average CRPS by 5.74\% relative to the strongest competing baseline. Its advantage is more pronounced for distributional quality than for median accuracy: SPECTRA obtains the best $\rho_{90}$ in 14 settings, with an average improvement of 7.27\%, compared with 1.78\% for $\rho_{50}$. These results indicate that SPECTRA improves not only the forecast center but, more importantly, the overall predictive distribution and upper-tail risk estimation.

The metric-dependent results further show that median accuracy and probabilistic quality are related but not equivalent. PatchTST, TimeXer, TiDE, Autoformer, and DLinear remain competitive in several $\rho_{50}$ cases, particularly for smooth or strongly autocorrelated series, but their gains are less consistent for CRPS and $\rho_{90}$. In contrast, SPECTRA maintains stronger performance across heterogeneous energy domains. This consistency is supported by its component-aware design: MTPD separates the trend-periodic structure from high-frequency residuals, STSSE refines structured dependencies, ECS aligns weather, calendar, and market context with both streams, and SBE converts residual variations into ordered quantile boundaries.

Fig.~\ref{fig:result_visualization} provides representative visual evidence. In the upper-row plots, the median forecasts track the regular trajectories, while the 80\% prediction intervals generally cover the observed future values and widen around ramps, spikes, and volatile transitions. The lower-row decomposition further shows that the deterministic stream captures smooth operating structure, whereas the residual stream corrects uncertain local variations. Although PatchTST or DLinear performs best on a few metrics, SPECTRA's broader gains in CRPS and $\rho_{90}$ support the joint modeling of temporal structure, exogenous context, and residual uncertainty.

\subsection{Ablation Study}
\label{sec:exp_ablation}

\begin{table}[t]
\centering
\caption{Ablation Study on ECL and OPS Datasets}
\label{tab:model_variants}
\resizebox{\columnwidth}{!}{%
\begin{tabular}{l|ccccc}
\toprule
\multirow{2}{*}{Model Variant} & \multirow{2}{*}{ECL} & \multicolumn{4}{c}{OPS} \\
\cmidrule(lr){3-6}
& & Load & Price & Solar & Wind \\
\midrule
    \textbf{SPECTRA}     & \textbf{0.156} & \textbf{0.118} & 0.348 & \textbf{0.131} & \textbf{0.442} \\
    \midrule
    w/o ECS              & 0.163 (+4.5\%) & 0.129 (+9.3\%) & 0.349 (+0.3\%) & 0.139 (+6.1\%) & 0.448 (+1.4\%) \\
    w/o STSSE            & 0.171 (+9.6\%) & 0.134 (+13.6\%) & 0.356 (+2.3\%) & 0.143 (+9.2\%) & 0.450 (+1.8\%) \\
    w/o High Freq        & 0.174 (+11.5\%) & 0.138 (+16.9\%) & 0.359 (+3.2\%) & 0.141 (+7.6\%) & 0.468 (+5.9\%) \\
    w/o Wavelet          & 0.159 (+1.9\%) & 0.123 (+4.2\%) & 0.350 (+0.6\%) & 0.135 (+3.1\%) & 0.449 (+1.6\%) \\
    \midrule
    Direct Quantile head & 0.157 (+0.6\%) & 0.126 (+6.8\%) & \textbf{0.346} (-0.6\%) & 0.136 (+3.8\%) & 0.448 (+1.4\%) \\
    Moving average       & 0.162 (+3.8\%) & 0.131 (+11.0\%) & 0.352 (+1.1\%) & 0.140 (+6.9\%) & 0.450 (+1.8\%) \\
\bottomrule
\end{tabular}%
}
\end{table}

To assess the contribution of each design choice, Table~\ref{tab:model_variants} reports the CRPS of ablation variants on ECL and OPS. The full SPECTRA performs best on ECL, OPS-Load, OPS-Solar, and OPS-Wind, while remaining competitive on OPS-Price, demonstrating consistent cross-task gains. Removing the high-frequency residual stream causes the largest degradation, increasing CRPS by 11.5\% on ECL and 16.9\% on OPS-Load, which confirms its importance for uncertainty boundary estimation. Removing STSSE also yields substantial drops, particularly on ECL and OPS-Load, supporting its role in refining structured dependencies.

The remaining variants clarify the contributions of individual modules. Removing ECS mainly degrades OPS-Load and OPS-Solar, consistent with the importance of calendar and weather context in these tasks. Replacing MTPD with moving-average decomposition consistently reduces performance, indicating the advantage of input-adaptive spectral separation over fixed temporal windows. The wavelet branch provides smaller but stable gains through complementary multi-resolution modeling. Although the direct quantile head slightly improves OPS-Price, it performs worse on most other tasks, suggesting weaker robustness across heterogeneous energy domains.

\begin{figure*}[t!]
    \centering
    \includegraphics[width=\textwidth]{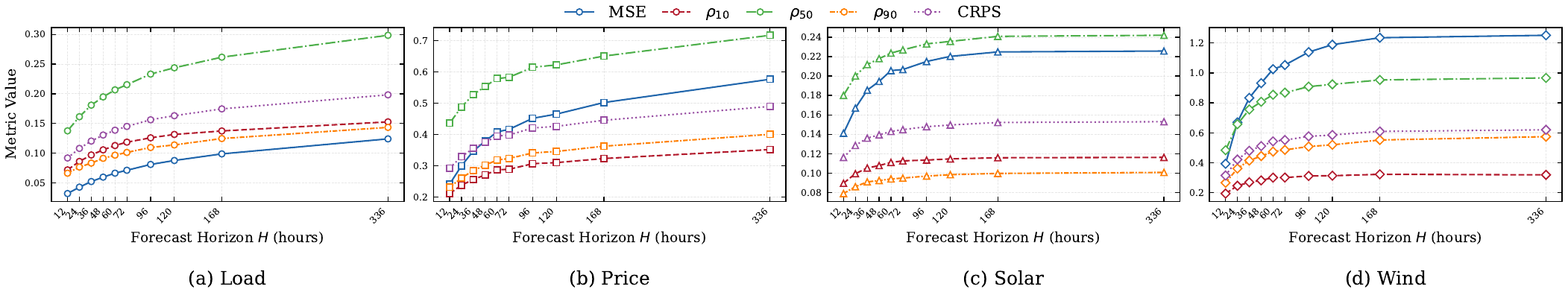}
    \caption{Impact of prediction length on forecast accuracy with the input length fixed as $L=168$.}
    \label{fig:longforecasting}
\end{figure*}

\begin{table}[t!]
\centering
\caption{Case Study Comparison on NE (Solar and Wind) Dataset}
\label{tab:case_study}
\resizebox{0.9\columnwidth}{!}{%
\begin{tabular}{ll|cccc}
\toprule
\multirow{2}{*}{Dataset} & \multirow{2}{*}{Metric} & \multicolumn{4}{c}{Model} \\
\cmidrule(lr){3-6}
& & PatchTST & TimeXer & TADNet & \textbf{SPECTRA} \\
\midrule
\multirow{2}{*}{Solar} & MSE  & \textbf{0.169} & 0.175 & 0.250 & 0.170 \\
                       & CRPS & 0.189 & 0.190 & 0.246 & \textbf{0.184} \\
\midrule
\multirow{2}{*}{Wind}  & MSE  & 0.220 & 0.219 & 0.245 & \textbf{0.216} \\
                       & CRPS & 0.218 & 0.214 & 0.246 & \textbf{0.212} \\
\bottomrule
\end{tabular}
}
\end{table}

\subsection{Case Study and Model Analysis}
\label{sec:exp_case_study}
Beyond average benchmark scores, we further examine whether SPECTRA generalizes to a recent renewable dataset, remains stable as the forecast horizon increases, and provides a favorable accuracy-efficiency trade-off.

\subsubsection{Case Study}
To assess generalization beyond standard public benchmarks, we conduct a case study on the NewEnergy2025 (NE) dataset, a recent renewable forecasting competition dataset with multi-site photovoltaic and wind generation. For each solar or wind site, the target is generation power, and the exogenous inputs are formed by concatenating meteorological variables from three nearby weather stations.

\textbf{Dataset Description:} For both solar and wind tasks, the look-back length and prediction horizon are set to $L=96$ and $H=96$, respectively. Since each task contains multiple sites, models are trained and evaluated on each site, and the average performance over all corresponding sites is reported. This setting evaluates cross-plant forecasting ability rather than performance on a selected station.

\textbf{Results:} Table~\ref{tab:case_study} presents the results. For solar forecasting, SPECTRA achieves the best CRPS of 0.184, with an MSE slightly higher than PatchTST, indicating more reliable predictive distributions while maintaining near-top point accuracy. For wind forecasting, SPECTRA obtains the lowest MSE and CRPS, showing robustness under stronger meteorological randomness. Overall, the NE case study supports the generalizability of SPECTRA to modern renewable generation forecasting and validates the effectiveness of exogenous context alignment and uncertainty-aware modeling.

\subsubsection{Long-Term Forecasting Performance}
\label{sec:exp_long_term}

We evaluate SPECTRA under prediction horizons from $H=12$ to $336$ with $L=168$. As shown in Fig.~\ref{fig:longforecasting}, deterministic and probabilistic errors generally increase with the horizon. Load exhibits a gradual degradation due to its regular temporal patterns, whereas price deteriorates more rapidly because of market volatility. Solar and wind errors rise mainly at short-to-medium horizons and then tend to stabilize, indicating that long-term forecasting difficulty is strongly target-dependent.

\subsubsection{Efficiency Analysis}
\label{sec:exp_model_analysis}
\begin{figure}[t!]
    \centering
    \includegraphics[width=0.4\textwidth]{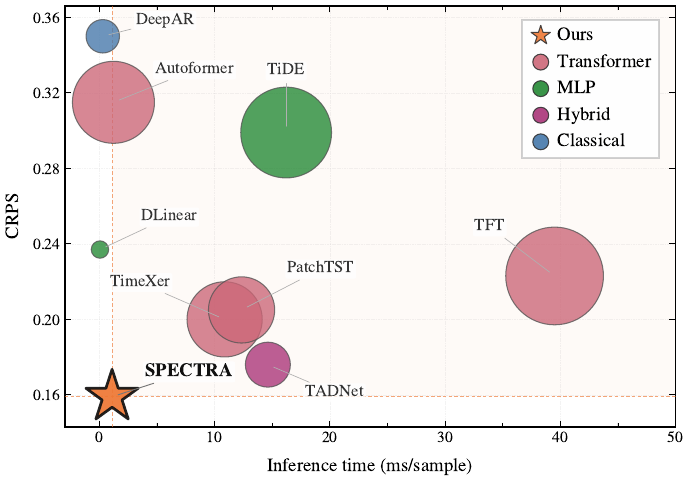}
    \caption{Efficiency comparison of different forecasting models on the NE dataset. The x-axis denotes inference time, the y-axis denotes CRPS, and the bubble area represents the number of parameters.}
    \label{fig:efficiency}
\end{figure}

Finally, we evaluate whether the probabilistic gain is obtained with acceptable overhead. Fig.~\ref{fig:efficiency} compares CRPS, inference time, and model size. SPECTRA is located near the lower-left region with a moderate bubble size, indicating low CRPS, low inference cost, and a compact parameter scale. Its efficiency mainly comes from the state-space backbone, which avoids the quadratic cost of full self-attention. Compared with DLinear, SPECTRA introduces moderate overhead but substantially improves probabilistic accuracy. Compared with attention-heavy models such as TFT, PatchTST, TimeXer, and Autoformer, it provides a more favorable accuracy-efficiency trade-off.

%% file: sections/5_conclusion.tex
\section{Conclusion}
\label{sec:conclusion}

This paper proposed SPECTRA, a state-space architecture that integrates adaptive temporal-frequency decomposition, exogenous-context alignment, state-space refinement, and ordered quantile estimation for general probabilistic energy forecasting. By separating trend-periodic structures from high-frequency residual variations, SPECTRA models the forecast center and uncertainty-bearing fluctuations through complementary pathways. Experiments across load, price, solar, and wind tasks demonstrate consistent improvements in CRPS and upper-tail quantile risk. Ablation studies verify the contributions of residual modeling, exogenous-context alignment, and spectral-temporal refinement, while the NE case study and efficiency analysis further demonstrate its generalization ability and favorable accuracy-efficiency trade-off. Future work will investigate robust forecasting under imperfect exogenous information, spatial-temporal probabilistic modeling, and the integration of large language models or time-series foundation models for cross-task transfer and decision-oriented interpretation.